\documentclass[10pt,twocolumn,letterpaper]{article}

\usepackage{cvpr}
\usepackage{times}
\usepackage{epsfig}
\usepackage{graphicx}
\usepackage{amsmath}
\usepackage{amssymb}
\usepackage{color,soul}
\usepackage{microtype}
\usepackage{textcomp}
\usepackage[table]{xcolor}
\usepackage[page]{appendix}
\usepackage{caption}
\usepackage{threeparttable}
\usepackage{tabularx}

\usepackage[pagebackref=true,breaklinks=true,letterpaper=true,colorlinks,bookmarks=false]{hyperref}

\cvprfinalcopy 

\def\cvprPaperID{10309} 

\definecolor{ForestGreen}{HTML}{228b22}

\ifcvprfinal\fi

\begin{document}

\title{xBD: A Dataset for Assessing Building Damage from Satellite Imagery}

\author{Ritwik Gupta\textsuperscript{1,2}\enspace Richard Hosfelt\textsuperscript{1,2}\enspace Sandra Sajeev\textsuperscript{1,2}\enspace Nirav Patel\textsuperscript{3,4} \enspace Bryce Goodman\textsuperscript{3,4}\enspace \\ Jigar Doshi\textsuperscript{5}\enspace Eric Heim\textsuperscript{1,2}\enspace Howie Choset\textsuperscript{1}\enspace Matthew Gaston\textsuperscript{1,2}\\
	\textsuperscript{1}Carnegie Mellon University\enspace \textsuperscript{2}Software Engineering Institute\enspace \textsuperscript{3}Defense Innovation Unit\\ \textsuperscript{4}Department of Defense\enspace \textsuperscript{5}CrowdAI, Inc.}

\maketitle

\begin{abstract}
We present xBD, a new, large-scale dataset for the advancement of change detection and building damage assessment for humanitarian assistance and disaster recovery research.
Natural disaster response requires an accurate understanding of damaged buildings in an affected region.
Current response strategies require in-person damage assessments within 24-48 hours of a disaster.
Massive potential exists for using aerial imagery combined with computer vision algorithms to assess damage and reduce the potential danger to human life.
In collaboration with multiple disaster response agencies, xBD provides pre- and post-event satellite imagery across a variety of disaster events with building polygons, ordinal labels of damage level, and corresponding satellite metadata.
Furthermore, the dataset contains bounding boxes and labels for environmental factors such as fire, water, and smoke.
xBD is the largest building damage assessment dataset to date, containing 850,736 building annotations across 45,362 km\textsuperscript{2} of imagery.
\end{abstract}

\vspace{-0.5cm}
\section{Introduction}
Resource allocation, aid routing,  rescue and recovery, and many other tasks in the world of humanitarian assistance and disaster response (HADR) can be made more efficient by using algorithms that  adapt to dynamic environments.
To accomplish these tasks in the context of natural disaster relief, it is necessary to understand the amount and extent of damaged buildings in the area.
Collecting this data is often dangerous, as it requires people on the ground to directly assess damage during or immediately after a disaster.
With the increased availability of satellite imagery, this task has the potential to not only be done remotely, but also automatically by applying powerful computer vision algorithms.

\begin{figure}[ht]
	\begin{center}
		\includegraphics[width=1\linewidth]{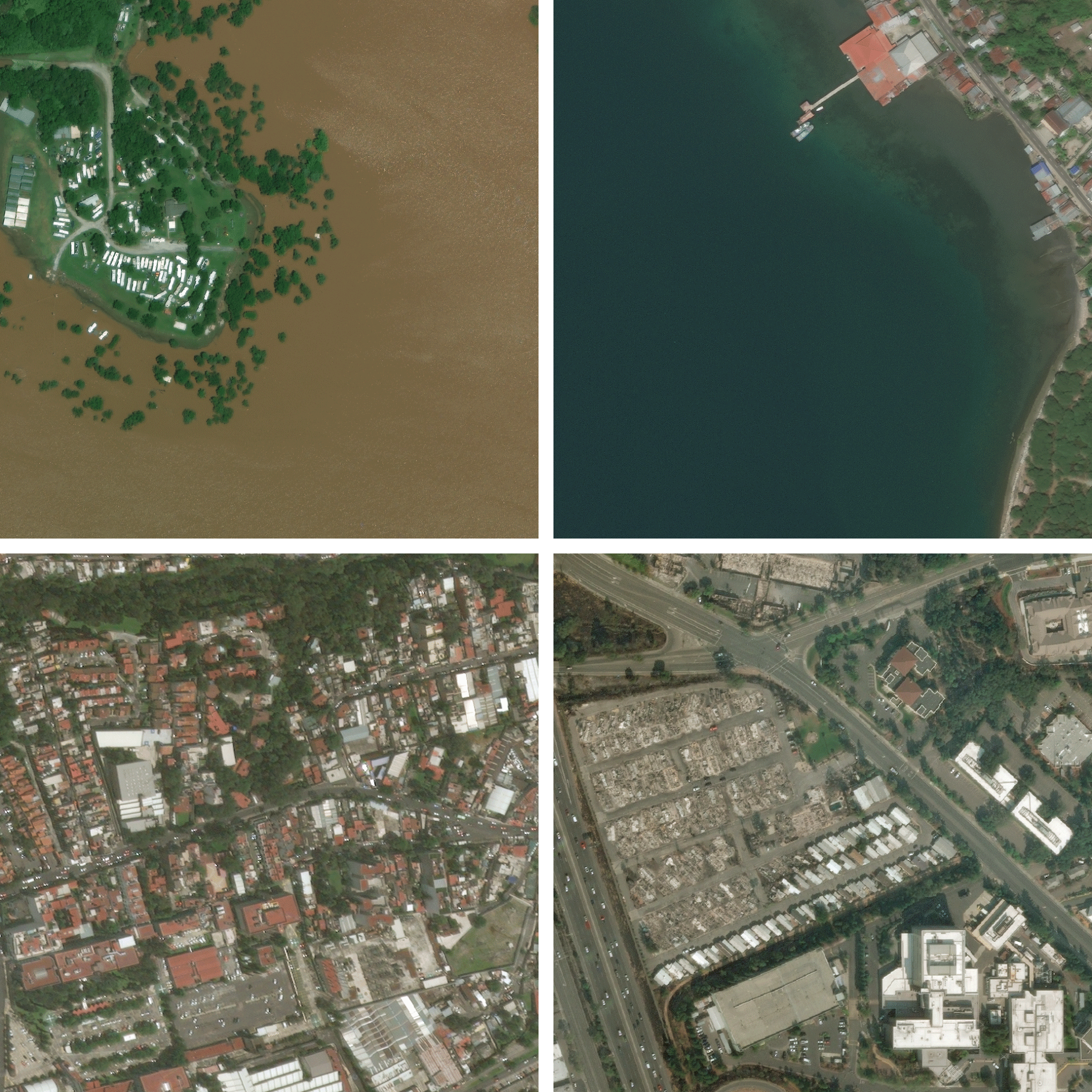}
	\end{center}
	\caption{From top left (clockwise): Hurricane Harvey; Palu Tsunami; Mexico City Earthquake; Santa Rosa Fire. Imagery from DigitalGlobe.}
	\vspace{-0.5cm}
	\label{fig:teaser}
\end{figure}
Currently,  adequate satellite imagery that addresses building damage is  not generally available.
To model the complex and dynamic nature of damage, imagery containing many types of damage must be available in large quantities.
To this end, we introduce xBD: a large-scale dataset of satellite imagery.  xBD covers a diverse set of disasters and geographical locations with over 800,000 building annotations across over 45,000 km\textsuperscript{2} of imagery.
To curate this dataset, we worked with disaster response experts from around the world who specialized in various disaster types to create an annotation scale that accurately represented real-world damage conditions.
Furthermore, we created and executed a rigorous, repeatable, and verifiable annotation process that ensured high-quality annotations with supporting quality control from experts.

The xBD dataset is used by the xView 2 prize challenge, which aims to spur the creation of accurate and efficient machine learning models that assess building damage from pre- and post-disaster satellite imagery.

In the rest of this paper, we go into details of the xBD dataset.
We begin by covering the requirements of the dataset, the annotation scale, the collection process, dataset statistics, the baseline model created for the xView 2 challenge, and potential use cases for the dataset beyond building damage classification.
This work builds on top of \cite{guptaCreatingXBDDataset2019} and presents a finalized report on the entire xBD dataset.

\begin{figure*}[!h]
	\begin{center}
		\includegraphics[width=1\linewidth]{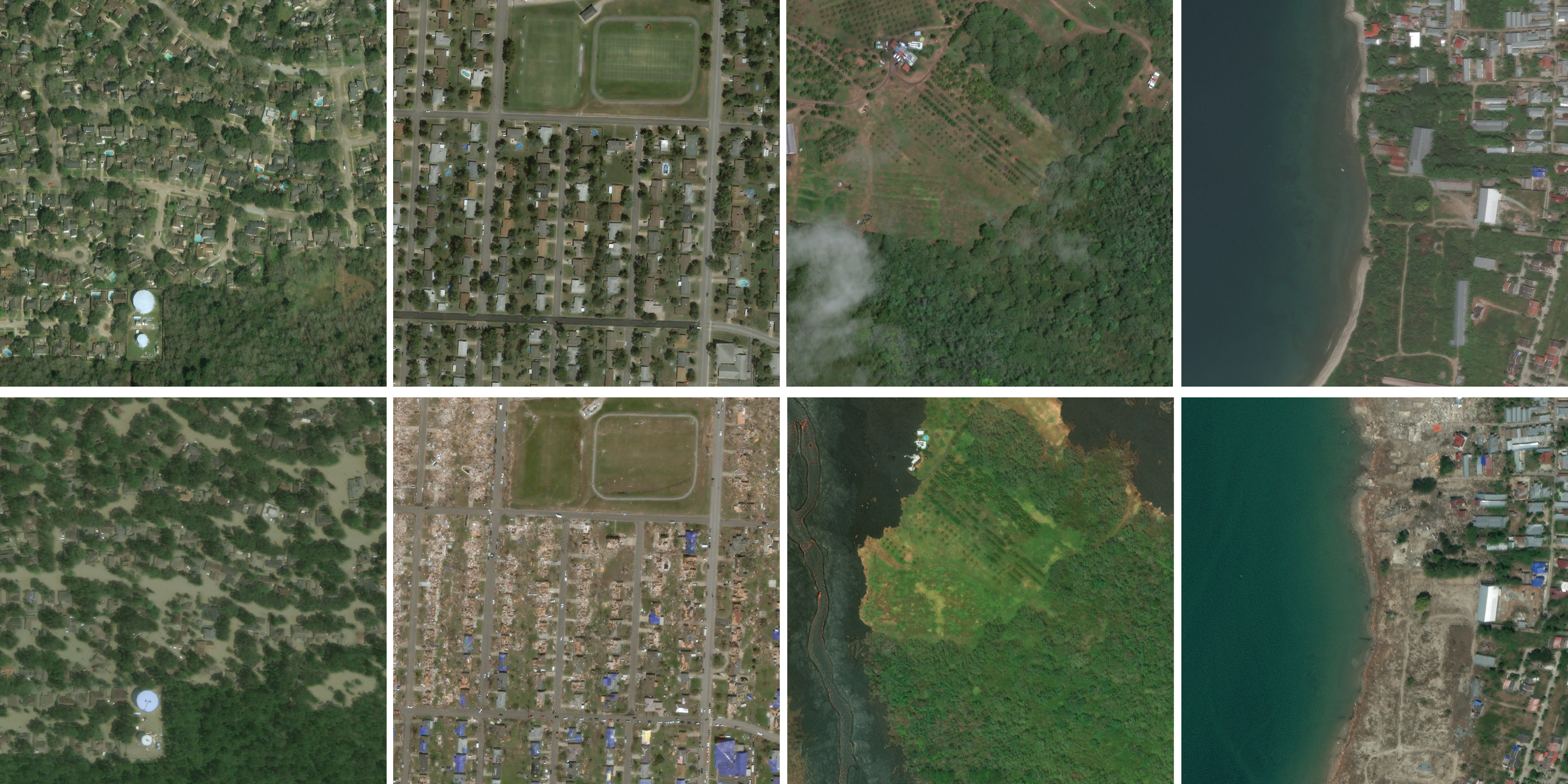}
	\end{center}
	\caption{Pre-disaster imagery (top)  and post-disaster imagery (bottom). From left to right: Hurricane Harvey; Joplin tornado; Lower Puna volcanic eruption; Sunda Strait tsunami. Imagery from DigitalGlobe.}
	\label{fig:teaser}
\end{figure*}

\section{Related Work}\label{sec:relatedwork}
Many factors  increase the complexity of damage assessment from satellite imagery.
The types and locations of disasters have a significant impact on the resulting damage.
Intuitively, damage after a flood in the United States will look substantially different from the aftermath of an earthquake in Nepal.
In this section, we will explore the attributes of existing satellite imagery datasets for damage assessment.

\subsection{Existing Data Sources}
Current satellite imagery datasets are often limited to single disaster types and do not have common criteria for assessing damage assessment ~\cite{fujitaDamageDetectionAerial2017, chenBenchmarkDatasetAutomatic2018, ColombiaBuildingDamage2017, foulser-piggottUseRemoteSensing2012}.
As a result, there is no meaningful way to compare damage assessments between a tsunami in Japan ~\cite{fujitaDamageDetectionAerial2017} and a mudslide in Columbia ~\cite{ColombiaBuildingDamage2017}.

After many discussions with disaster response experts, it was made clear that damage exists on a continuum. However, because the personnel and time available for imagery analysis are often limited, a damage scale that is more developed than a simple binary ``damaged"/``undamaged" option ~\cite{fujitaDamageDetectionAerial2017} has not been explored.

An issue that many HADR imagery analysts face is the presence of haze or mild occlusion in electro-optical (EO) imagery.
As a result, they sometimes must resort to lower resolution synthetic aperture radar (SAR) imagery, which may not always be available for the event under analysis.
Cloud occlusion in EO imagery is a well-known problem that has been addressed via many traditional computer vision and deep learning techniques ~\cite{mitchellFilteringRemoveCloud1977, baiCloudDetectionHighResolution2016, zhuRemoteSensingImaging2017}.

Analysts regularly use contextual information in satellite imagery to improve their damage assessments. These contexts often include environmental factors such as the presence of water, fire, smoke, and more.  Relatively few datasets ~\cite{demirDeepGlobe2018Challenge2018, giglioAnalysisDailyMonthly2013}  provide information of this granularity.

\subsection{Assessment Scales}
Most HADR organizations need to assess many types of building damage across different regions and disaster types.
A number of damage assessment scales are available in existing disaster response literature. However, these scales are largely limited to a single disaster type or are specifically focused on in-person assessment ~\cite{jiIdentifyingCollapsedBuildings2018, thiekenMethodsEvaluationDirect, okadaClassificationsStructuralTypes1999, friedlandResidentialBuildingDamage2009}. 
~\cite{jiIdentifyingCollapsedBuildings2018} developed a damage scale for building damage assessment from satellite imagery, but the scope was narrow and focused on earthquakes.
Having multiple, disjoint scales when responding to a disaster results in a large cognitive burden for HADR organizations.

HAZUS from FEMA is a robust scale for multi-disaster damage classification, including earthquake, floods, and hurricanes ~\cite{federalemergencymanagementagencyHazusHurricaneModel2018}.
However, some of the criteria for assessment require an analyst to be 
physically present at the disaster site.
For instance, the HAZUS Hurricane model considers the roof, windows, and wall structure condition to determine damage in a residential setting ~\cite{federalemergencymanagementagencyHazusHurricaneModel2018}.
Some of these attributes, like wall structure and windows, cannot easily be determined from satellite imagery. 

FEMA also created the FEMA Damage Assessment Operations Manual, which is specifically intended for qualitative damage assessment ~\cite{federalemergencymanagementagencyDamageAssessmentOperations2016}.
However, like HAZUS, it is primarily designed for in-person assessment. 

\section{Requirements for the xBD Dataset}\label{sec:requirements}
xBD was created to fulfill a specific need in the world of HADR.
A set of requirements were created to guide the collection, labeling, quality control, and dissemination of the data.
However,  to make the dataset useful for a general set of problems, certain requirements were tweaked or modified to fit the needs of the larger research community without compromising the original dataset need.

\subsection{Multiple Levels of Damage}
After discussions with disaster response experts from CAL FIRE and the California Air National Guard, it was clear that agencies did not currently have the capacity to classify multiple levels of damage.
Many analysis centers simply label buildings as ``damaged" or ``undamaged"  to reduce the amount of expert man-hours needed for assessment, even though it was clear that damage is not a binary status.
Discerning between multiple levels of damage is a critical mission need, therefore xBD needed to represent a continuum of damage.

\subsection{Image Resolution}
Differences between levels of damage are often visually minute.
To facilitate the labeling of these types of damage,  supporting imagery must be of high fidelity and have enough discerning visual cues.
We targeted satellite imagery to be below a 0.8 meter ground sample distance (GSD) mark to fulfill this requirement.

\subsection{Diversity of Disasters}
One goal of the xView 2 prize challenge is to output models that are widely applicable across a large number of disasters.
This will enable multiple disaster response agencies to potentially reduce their workload by using one model with a known deployment cycle.
xBD would need to be representative of multiple disaster types and not simply the ones for which a large amount of data was commonly available.

\begin{figure*}[ht]
	\begin{center}
		\includegraphics[width=1\linewidth]{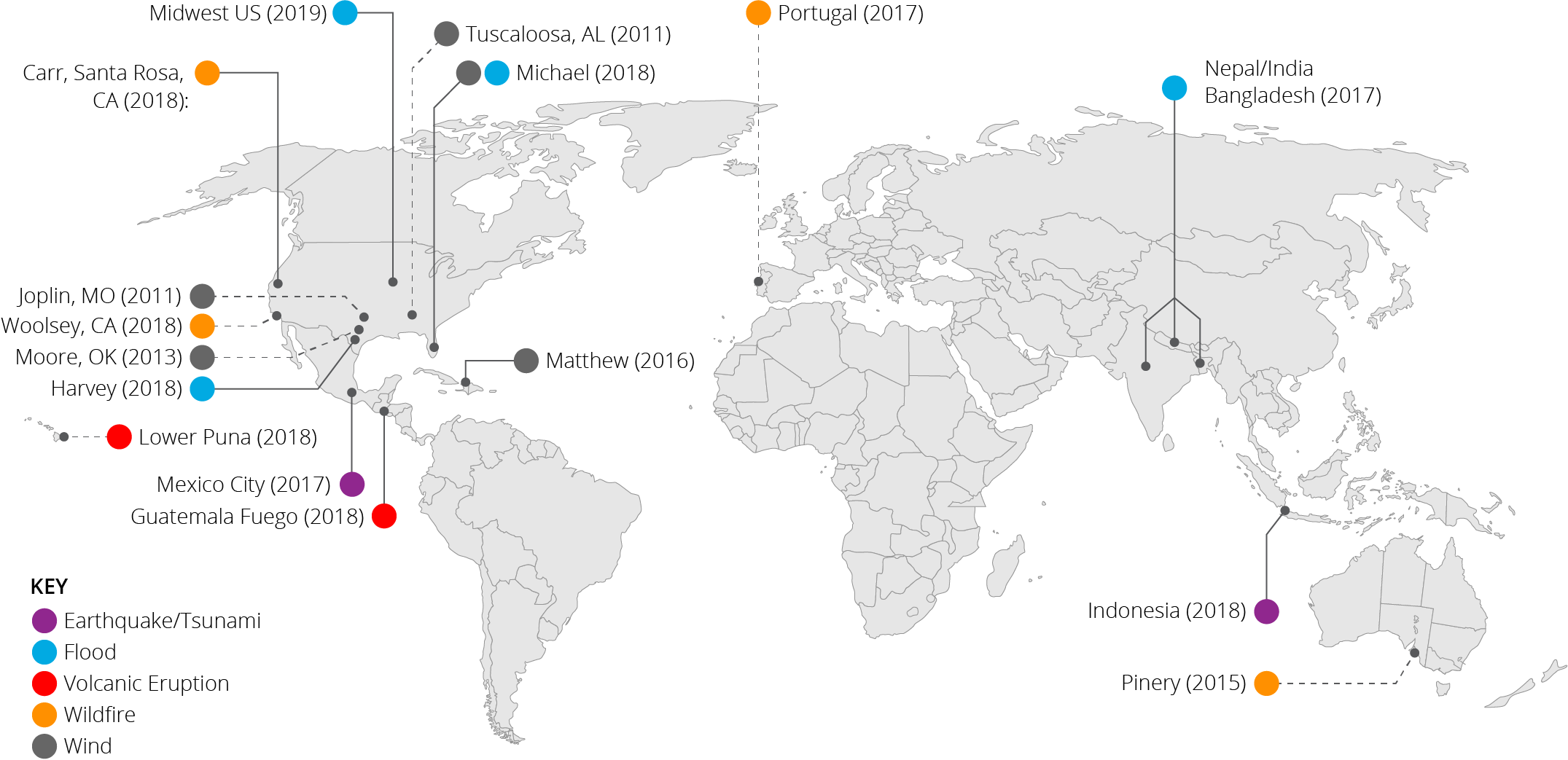}
	\end{center}
	\caption{Disaster types and disasters represented in xBD around the world.}
	\label{fig:teaser}
\end{figure*}

\subsection{Wide Diversity of Buildings}
Building damage looks different based on the type of building as well as the locale in which the building exists.
To properly represent this variance in building damage, xBD sampled imagery from various regions of the world that accounts for varying building sizes, densities, and more.

\subsection{Negative imagery}
Examples of imagery with no damage or no damaged buildings are critical for learning the difference between different levels of disaster. 
Negative imagery therefore should comprise a sizable percentage of the dataset.

In the subsequent sections, we highlight how xBD was specifically designed to satisfy each of these requirements.

\section{Joint Damage Scale}
\begin{figure}[ht]
	\begin{center}
		\includegraphics[width=1\linewidth]{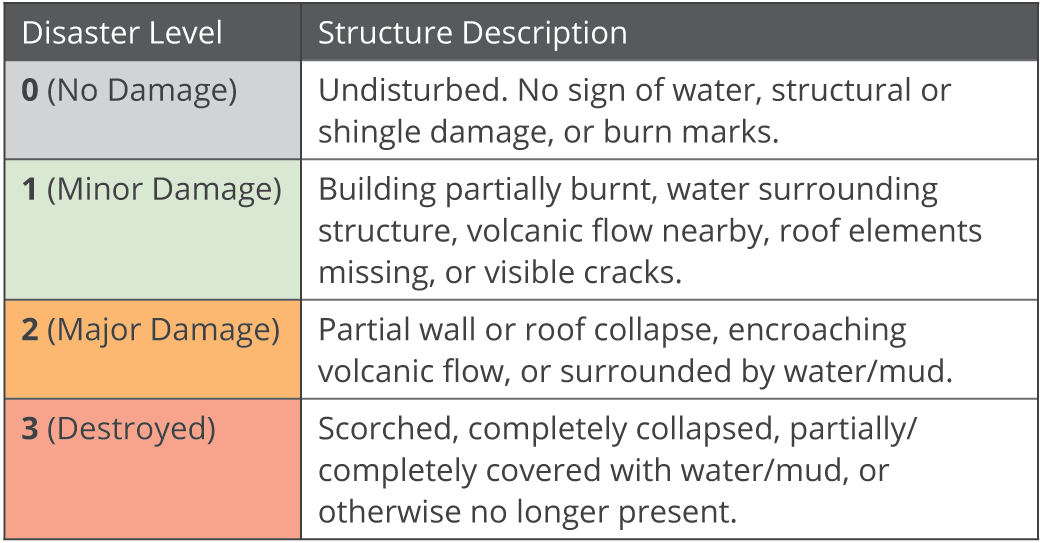}
	\end{center}
	\caption{Joint Damage Scale descriptions on a four-level granularity scheme.}
	\vspace{-0.5cm}
	\label{table:joint.damage.scale}
\end{figure}

To assess damage from different types of disasters, disaster response agencies currently have to use a variety of different scales.
This increases the cognitive burden on the analysts and limits the transfer capability of labels across datasets.
We present the Joint Damage Scale (Table \ref{table:joint.damage.scale}) for the assessment of building damage from satellite imagery across multiple disaster types.
The Joint Damage Scale is created with insight from NASA, CAL FIRE, FEMA, and the California Air National Guard.
Furthermore, the Joint Damage Scale is grounded in existing literature, such as HAZUS~\cite{federalemergencymanagementagencyHazusHurricaneModel2018}, FEMA's Damage Assessment Operations Manual~\cite{federalemergencymanagementagencyDamageAssessmentOperations2016}, the Kelman scale~\cite{kelmanPhysicalFloodVulnerability2002}, and the EMS-98 scale~\cite{grunthalEMS98EuropeanMacroseismic1998}.

This scale is not meant as an authoritative damage assessment rating. However, it does serve as a first attempt to create a unified assessment scale for building damage in satellite imagery across multiple disaster types, structure categories, and geographical locations.

The Joint Damage Scale ranges from no damage (0) to destroyed (3).
After multiple iterations with analysts, this level of granularity was chosen as an ideal trade-off between utility and ease of annotation.
The descriptions of each damage level have been generalized to handle the wide variety of disasters present in xBD, which can result in some amount of label noise.
Although such nuance is not ideal from an objective assessment standpoint, it allows analysts to gracefully handle tough cases that fall between classification boundaries.

Due to the limitations presented by the modality of satellite imagery, such as resolution, azimuth, and smear~\cite{wahballahSmearEffectHighresolution2018}, this scale presents a best-effort trade-off between operational relevance and technical correctness. Thus, the Joint Damage Scale cannot accommodate the degree of precision that a scale meant for in-person, human assessment provides.

\section{Dataset Collection}
\subsection{Imagery Source}
Imagery for xBD was sourced from the Maxar/DigitalGlobe Open Data Program\footnote{https://www.digitalglobe.com/ecosystem/open-data}, which releases imagery for sudden onset major crisis events.
Specifically, the Open Data Program was chosen for its availability of high-resolution imagery from many disparate regions of the world.
Of the multitude of open data events that are currently available in the Open Data Program's catalog, we selected 11 disaster events from an initial list of 19.
These 11 events are dubbed ``Tier 1" events.
In addition to the Tier 1 events, we identified 8 additional events that were not a part of the Open Data Program, which we dubbed ``Tier 3" events.
We partnered with Maxar and the National Geospatial-Intelligence Agency to activate specific areas of interest (AOIs) from Tier 3 for inclusion in the Open Data Program.
A list of these events is provided in Table \ref{table:disasters}. All imagery is available in three-band RGB formats.

\begin{table*}
	\small
	\begin{tabularx}{2\columnwidth}{X|X|c|c}
		\hline
		\textbf{Disaster Event Name} & \textbf{Event Dates} & \textbf{Tier} & \textbf{Environmental Factors} \\
		\hline 
		Guatemala Fuego Volcano Eruption & Jun 3, 2018 & 1 & Yes \\ 
		\hline 
		Hurricane Michael & Oct 7-16, 2018 & 1 & No \\ 
		\hline 
		Santa Rosa Wildfires & Oct 8-31, 2017 & 1 & Yes\\ 
		\hline 
		Hurricane Florence & Sep 10-19, 2018 & 1 & Yes \\ 
		\hline 
		Midwest US Floods & Jan 3 - May 31, 2019 & 1 & Yes \\ 
		\hline 
		Indonesia Tsunami & Sep 18, 2018 & 1 & Yes \\ 
		\hline 
		Carr Wildfire & Jul 23 - Aug 30, 2018 & 1 & No \\ 
		\hline 
		Hurricane Harvey & Aug 17 - Sep 2, 2017 & 1 & No \\ 
		\hline 
		Mexico City Earthquake & Sep 19, 2017 & 1 & No \\ 
		\hline 
		Hurricane Matthew & Sep 28 - Oct 10, 2016 & 1 & No \\ 
		\hline 
		Monsoon in Nepal, India, Bangladesh & Jul - Sep, 2017 & 1 & Yes \\ 
		\hline
		Moore, OK Tornado & May 20, 2013 & 3 & No \\ 
		\hline
		Tuscaloosa, AL Tornado & Apr 27, 2011 & 3 & No \\ 
		\hline
		Sunda Strait Tsunami & Dec 22, 2018 & 3 & No \\ 
		\hline
		Lower Puna Volcanic Eruption & May 23 - Aug 14, 2018 & 3 & Yes \\ 
		\hline
		Joplin, MO Tornado & May 22, 2011 & 3 & No \\ 
		\hline
		Woolsey Fire & Nov 9-28, 2018 & 3 & No \\ 
		\hline
		Pinery Fire & Nov 25 - Dec 2, 2018 & 3 & No \\ 
		\hline
		Portugal Wildfires & Jun 17-24, 2017 & 3 & No \\ 
		\hline
	\end{tabularx}
	\caption{Disasters selected for xBD.}
	\label{table:disasters}
\end{table*}

\subsection{Annotation Process}
Annotation followed a multi-step process that  created polygons and damage classifications with a web-based annotation tool that was developed in-house by CrowdAI.

\vspace{-0.2cm}
\subsubsection{Triage}
Of the imagery provided by the Open Data Program, only a small amount contained actual instances of damage.
To reduce the amount of wasted annotation time, we manually reviewed all available imagery to identify AOIs that were usable for annotation.
However, each AOI purposefully included small amounts of buffer area (including the surrounding regions) to ensure the availability of negative imagery.

\vspace{-0.2cm}
\subsubsection{Imagery Matching and Polygon Annotation}
\begin{figure}[!h]
	\begin{center}
		\includegraphics[width=0.75\linewidth]{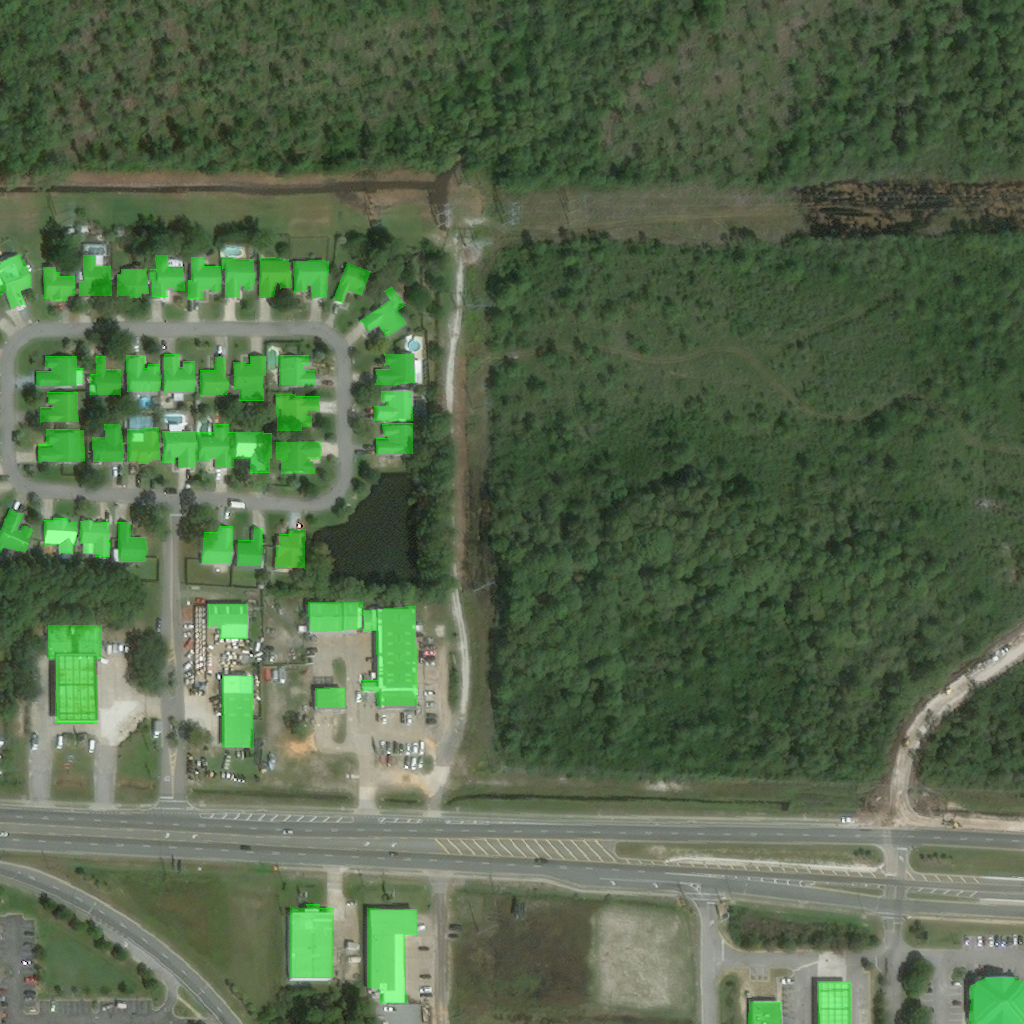}
	\end{center}
	\caption{Building polygons (shown in green) on imagery from Hurricane Michael (2018)}
	\label{fig:pre.polygons}
\end{figure}

Once AOIs containing damage are identified in the post-disaster imagery (or post-imagery), the equivalent AOI must be found and aligned in the pre-disaster imagery (or pre-imagery).
After all pairs of images were created, the pre-imagery was sent to annotators to draw polygons around visible building footprints. An example of these pre-imagery building footprints is shown in Figure \ref{fig:pre.polygons}

\vspace{-0.2cm}
\subsubsection{Post-Imagery Polygons Damage Classification}
The polygons that were extracted from the pre-imagery were overlaid onto their matching post-imagery pair.
This provided the ``ideal" building footprint before any damage occurs, since the footprint of a building may be dramatically altered during a disaster.
Each image was then sent to a group of annotators where the buildings were classified based on the Joint Damage Scale.
The first round consisted of multiple instances of annotation.
After the initial round of classification, each image went through a round of review to ensure consistency in annotation.
Finally, sets of annotations were randomly sampled and reviewed by experts from the California Air National Guard, NASA, and FEMA to  validate whether the damage classifications were accurate.
These experts found that approximately 2-3\% of the annotations had been mislabeled.
The mislabeled images were subsequently corrected by the experts themselves.

\vspace{-0.2cm}
\subsubsection{Environmental Factors}
In addition to annotating building damage, we provided polygons that represent ``environmental factors" visible in any image included in the dataset.
These environmental factors include smoke, fire, flood water, pyroclastic flow, and lava.
Flood water was interpreted as water visible in an area where it is not normally expected to  be present.
This means areas of ``known water," such as rivers, lakes, and ponds, were excluded.
Table \ref{table:disasters} shows which disasters included detectable environmental factors.

\subsection{Design Trade-Offs}
Due to various issues with edge cases in classifying damage, re-projection, and imagery quality, many design trade-offs were carefully constructed and followed throughout the annotation process.

\vspace{-0.2cm}
\subsubsection{Image Shifting}
As part of the data finalization process for each disaster event, we processed the post-disaster imagery to shift the image pixels slightly.
This was to account for re-projection issues when overlaying the building polygons from the pre-imagery onto the post-imagery.
Since the paired images were taken at different times (and sometimes with different sensors), small changes in off-nadir angle, sun elevation angle, and other considerations resulted in “polygon drift,” where the polygons were consistently misaligned by a fixed amount of pixels across an entire image.

After damage classification was complete, we sampled many random image tiles from the post-disaster CatID and calculated the average pixelwise shift by measuring the number of pixels between the edge of the polygon and the edge of the corresponding building side.
The number of pixels was then used to compute the shift in meters based on the ground sample distance of the image.
The corresponding UTM (Universal Transverse Mercator) shift was then applied to all post-disaster images within that disaster event uniformly.
Since this was a coordinate system-level translation, polygons were not shifted, nor was any change made to the pre-disaster imagery.

\vspace{-0.3cm}
\subsubsection{Building Polygons Not Present in Post-Disaster Imagery}
In some cases, buildings that are visually present in the post-disaster imagery  do not have a corresponding polygon.
The most common causes of this discrepancy were 1) the building did not exist in the pre-imagery (i.e., the building was constructed after the disaster); 2) the building did not meet the definition of a building  in the pre-imagery (e.g., the building was still under construction); or 3) the building was sufficiently occluded by clouds, haze, or vegetation that an accurate polygon could not be drawn.
Instances of discrepancies 1) and 2) were negligible and occurred mainly in datasets that spanned a large amount of time, such as the US Midwest Floods dataset.
In the annotation instructions for post-imagery, annotators were told to ignore buildings in the post-imagery that did not have polygons transferred over from pre-imagery.

\vspace{-0.3cm}
\subsubsection{Ground Sample Distance (GSD) Differences}
The GSD values associated with images of the same geographic region can vary due to a variety of factors.
However, these factors stem from the imagery itself, and not due to the data annotation process.
Furthermore, Maxar uses a proprietary pansharpening algorithm whose resulting GSD is not reported in the Open Data Program metadata.
The final imagery GSD can be estimated using various methods in the GDAL \cite{gdal/ogrcontributorsGDALOGRGeospatial2019} toolkit.

\section{Dataset Analysis}
The complete xBD dataset contains satellite images from 19 different natural disasters across 22,068 images and contains 850,736 building polygons.
The imagery covers a total of 45,361.79 km\textsuperscript{2}.

\subsection{Dataset Split}
xBD is provided in train, test, and holdout splits in a 80/10/10\% split ratio, respectively.
The specific split counts are provided in Table \ref{table:split.counts}.
Compared to traditional train/validation/test splits, this dataset splitting strategy is meant to facilitate the xView 2 challenge.
Participants can split the provided training dataset into a validation set.
The test set is meant to be used as a fixed evaluation set during the open leaderboard phase of the challenge.
The holdout set is purposefully not released during the duration of the challenge and is meant to be used as a private evaluation set to counter any challenge-specific gaming.
All dataset splits will be made available at the end of the xView 2 challenge.

\begin{table}[!h]
	\begin{tabularx}{\columnwidth}{X|X|X}
		\hline
		\textbf{Split} & \textbf{Images} & \textbf{Polygons} \\
		\hline 
		Train & 18,336 & 632,228 \\ 
		\hline 
		Test & 1,866 & 109,724 \\
		\hline
		Holdout & 1,866 & 108,784 \\
		\hline
	\end{tabularx}
	\caption{xBD data splits and their respective annotation counts.}
	\label{table:split.counts}
\end{table}

\subsection{Dataset Statistics}
\begin{figure}[!h]
	\begin{center}
		\includegraphics[width=0.80\linewidth]{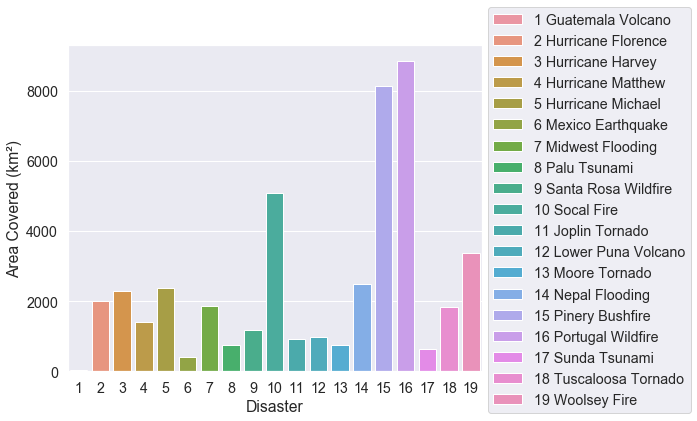}
	\end{center}
	\caption{Area of imagery (in km\textsuperscript{2}) per disaster event.}
	\label{fig:area.per.disaster}
\end{figure}

Disaster events are unevenly represented with regards to their total imagery area in xBD.
Figure \ref{fig:area.per.disaster} shows the distribution of areas covered from each disaster.

\begin{figure}[h]
	\begin{center}
		\includegraphics[width=0.99\linewidth]{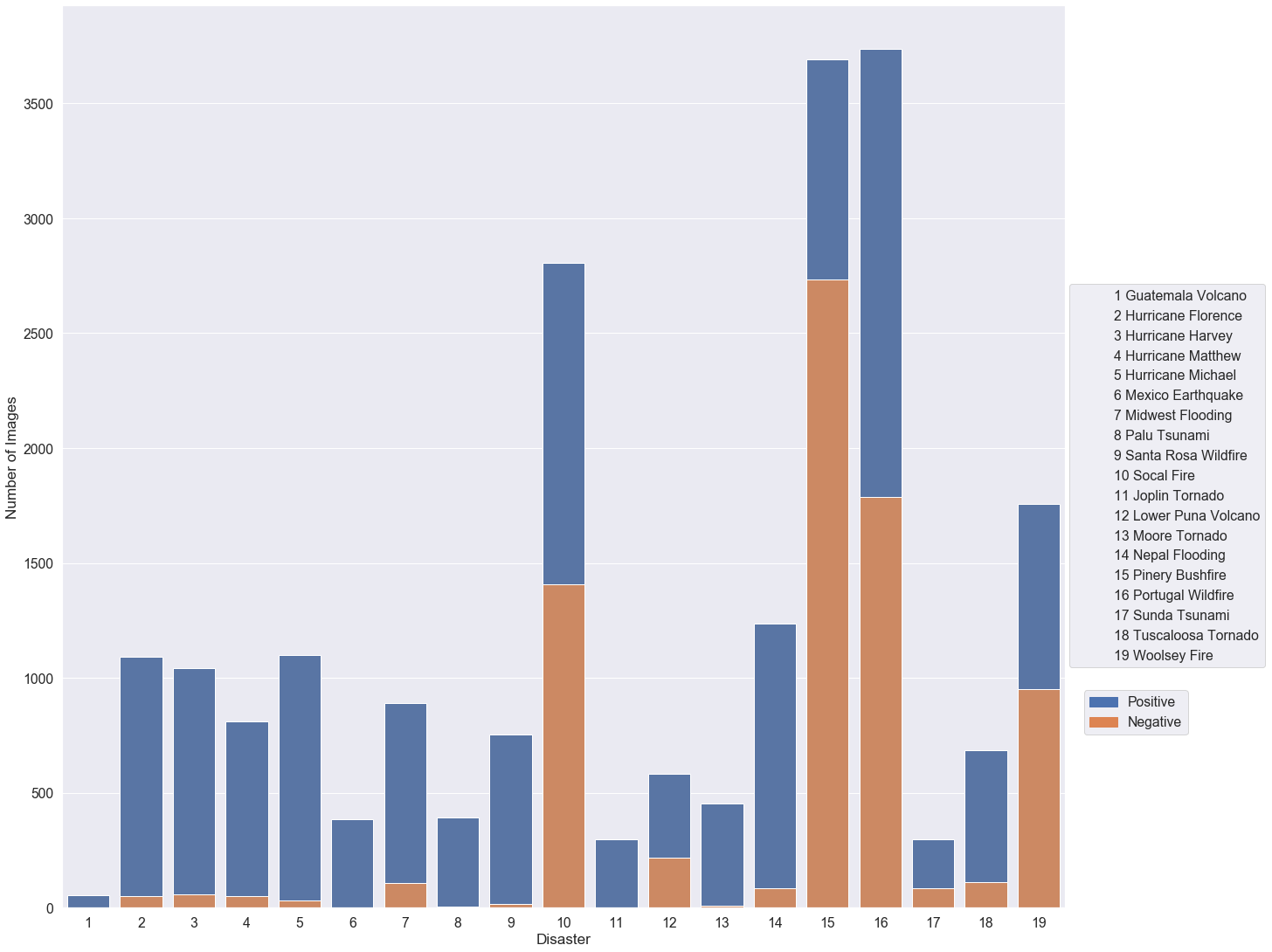}
	\end{center}
	\caption{Positive and negative imagery per disaster.}
	\label{fig:pos.neg}
\end{figure}

Interestingly, the amount of positive imagery that was contributed to the dataset does not directly correlate with the total area represented by each disaster.
Figure \ref{fig:pos.neg} shows the amount of positive and negative imagery in xBD as a function of individual disaster events.

\begin{figure}[h]
	\begin{center}
		\includegraphics[width=0.99\linewidth]{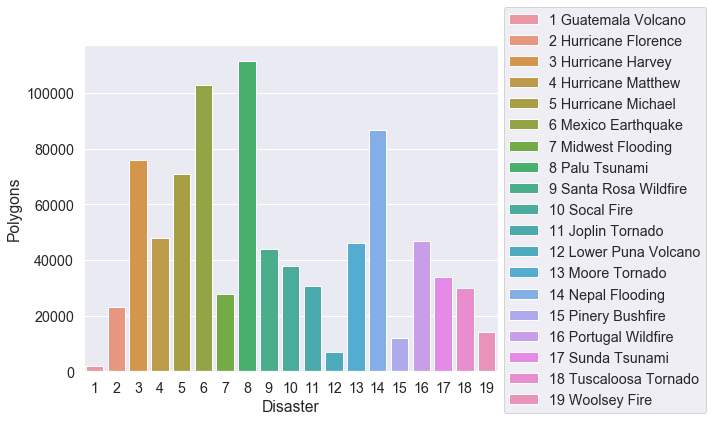}
	\end{center}
	\caption{Polygons in xBD per disaster event.}
	\label{fig:polygons.per.disaster.counts}
\end{figure}

Certain disasters are much more polygon dense than others.
Of note, the Mexico City earthquake and the Palu tsunami provide a large amount of polygons in comparison to their relatively low image areas.
Figure \ref{fig:polygons.per.disaster.counts} breaks this down per disaster.

\begin{figure}[h]
	\begin{center}
		\includegraphics[width=0.99\linewidth]{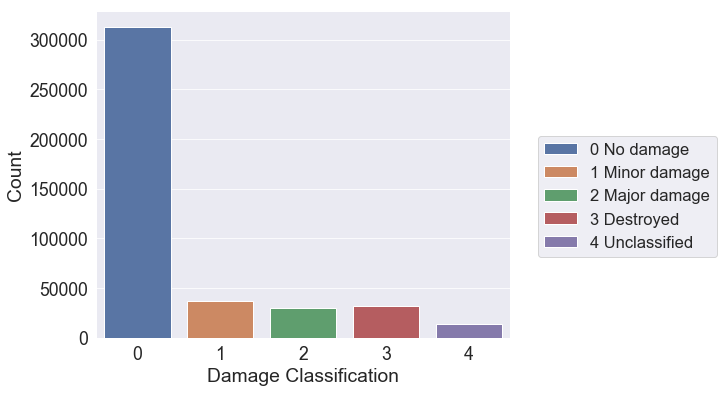}
	\end{center}
	\caption{Distribution of damage class labels.}
	\label{fig:label.counts}
\end{figure}

Finally, the distribution of damage classifications is highly skewed towards ``no damage," which had more than eight times the representation of the other classes.
``No damage," ``minor damage," ``major damage," and ``destroyed" are composed of 313,033, 36,860, 29,904, and 31,560 polygons respectively.
There are an additional 14,011 polygons labeled as ``unclassified." Figure \ref{fig:label.counts} breaks this difference down visually.

xBD presents an interesting challenge for the computer vision community because the distribution of labels and types of disasters is highly unbalanced.
Furthermore, the visual cues that differentiate the different levels of damage can be minute.
The varying availability of negative imagery for different types of disasters can also be a hurdle for localization models.





\section{Baseline Model}
Baseline localization and classification models were created to assess the difficulty of the xBD dataset and to serve as a starting point for the xView 2 challenge. All code is available at \href{https://github.com/DIUx-xView/xview2-baseline/}{https://github.com/DIUx-xView/xview2-baseline/}.

\subsection{Localization}
The localization model was  based on a SpaceNet\footnote{https://spacenetchallenge.github.io/} submission by Motoki Kimura\footnote{https://github.com/motokimura}, which featured an altered U-Net architecture \cite{ronnebergerUNetConvolutionalNetworks2015}.
We lightly modified this model to fit our dataset.
The model was trained on an eight GPU cluster for seven days. The model achieved an IoU of 0.97 and 0.66 for ``background" and ``building," respectively.

\subsection{Classification}
\begin{figure*}[!h]
	\begin{center}
		\includegraphics[height=2in]{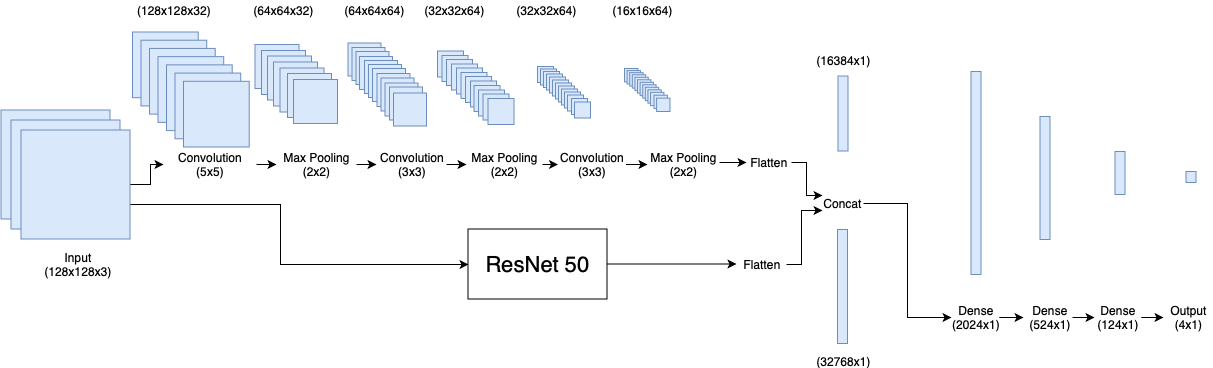}
	\end{center}
	\caption{Architecture of the baseline classification model. The input is fed into a pre-trained ResNet 50 as well as a shallow CNN. The outputs of each stream are concatenated and passed into dense layers for classification.}
	\label{fig:network.architecture}
\end{figure*}

The classification model is shown in Figure \ref{fig:network.architecture}. The ResNet50 is pre-trained on ImageNet \cite{dengImageNetLargeScaleHierarchical} whereas the smaller side network is initialized with random weights.
All convolutional layers use a ReLU activation.
The output is a one-hot encoded vector where each element represents the probability of an ordinal class.

The model uses an ordinal cross-entropy loss function.
Unlike traditional cross-entropy, ordinal cross-entropy penalizes relative to the distance between true and predicted ordinal class.
Since the difference between any two classes is not interchangeable, this loss function allows the model to better distinguish between the different levels of damage.

\subsection{Training}
The model was trained on 632,228 post-disaster image polygons with a batch size of 64.
The Adam optimizer with a learning rate of 0.001, $\beta_1$ of 0.9, and $\beta_2$ of 0.999 was used for training.
The network was trained for 100 epochs on 8 Nvidia GTX-1080 GPUs.

\subsection{Evaluation}
The weighted F1 score was the primary metric for evaluating the dataset.
Weighted F1 balances precision and recall in a harmonic mean and is well-suited for imbalanced datasets such as xBD.
Accuracy by itself is a flawed metric, since a classifier that predicted ``no damage" on all of the images would retain 75\% accuracy.

For the baseline model, we attained an overall weighted F1 score of 0.2654. Table \ref{table:baseline.f1} breaks down F1 score on a class-by-class basis.
``Major damage" instances were often classified as ``minor damage," resulting in a low recall.
We believe this result is due to a weak imbalanced training regimen, as greater separation between these classes can be made by a random classifier.
However, ``minor" and ``major" damage classes also have minute visual differences, and we expect most models to incur high confusion between these two classes.

\begin{table}[h]
	\centering 
	\begin{tabular}{l c c r} 
		\hline
		Damage Type & F1 Score & Precision & Recall
		\\ [0.5ex]
		\hline 
		No Damage & 0.6631 & 0.8770 & 0.5330 \\[-0.25ex]
		Minor Damage & 0.1435 & 0.1971 &  0.1128 \\[-0.25ex]
		Major Damage & 0.0094 & 0.7259 &  0.0047 \\[-0.25ex]
		Destroyed & 0.4657 & 0.5050 &  0.4321 \\[-0.25ex]
		\hline 
	\end{tabular}
	\caption{Baseline F1 Scores} 
	\label{table:baseline.f1}
\end{table}

\vspace{-0.5cm}
\section{Conclusion}
Natural disaster relief operations  require accurate assessments of building damage  to effectively allocate aid, personnel, and resources.
Currently, this is a dangerous and time-consuming manual task that must be carried out by human responders.
By recognizing that there are distinct visual patterns in damage that make this task automatable, xBD provides a large corpus of imagery and annotated polygons, along with an annotation scale. This will enable the creation of vision models that can automate the assessment of building damage and perform it remotely.
xBD contains a variety of disaster types across visually distinct regions of the world. This satisfies many real-world requirements as dictated by several disaster response agencies.

\vspace{1em}
\noindent \textbf{Acknowledgements:} The authors would like to thank Phillip SeLegue (California Department of Forestry and Fire Protection), Major Megan Stromberg and Major Michael Baird (California Air National Guard), Colonel Christopher Dinote (Joint Artificial Intelligence Center), and many others from the Indiana Air National Guard, United States Geological Survey, National Oceanic and Atmospheric Administration, National Aeronautics and Space Administration, Federal Emergency Management Agency, and the city of Walnut Creek, CA, for their subject matter expertise and assistance in creating xBD and the xView 2.0 challenge. We acknowledge the Maxar/DigitalGlobe Open Data program for the imagery used in this dataset.
Many thanks to Hollen Barmer, Nathan VanHoudnos, Diane Hosfelt, Abirami Kurinchi-Vendhan, and Courtney Rankin for reviewing this manuscript.
\vspace{\baselineskip}\linebreak
\noindent \textcopyright 2019 Carnegie Mellon University. This material is based upon work funded and supported by the Department of Defense under Contract No. FA8702-15-D-0002 with Carnegie Mellon University for the operation of the Software Engineering Institute, a federally funded research and development center.

\noindent [DISTRIBUTION STATEMENT A] This material has been approved for public release and unlimited distribution.  Please see Copyright notice for non-US Government use and distribution.
DM19-1227

{\small
	\bibliographystyle{ieee}
	\bibliography{xview_bib}
}
	
\end{document}